\documentclass[letterpaper, 10 pt, journal, twoside]{ieeetran} 
\usepackage[utf8]{inputenc}
\usepackage[T1]{fontenc}

\usepackage{graphics} 
\usepackage{epsfig} 
\usepackage{bm}
\usepackage{times} 
\usepackage{amsmath} 
\usepackage{amssymb}  
\usepackage{float}
\usepackage[space, compress]{cite}
\usepackage{mathtools}
\usepackage[ruled,vlined]{algorithm2e}
\usepackage{graphicx}
\usepackage{hyperref}
\usepackage{caption}
\usepackage{subcaption}
\usepackage{makecell}
\usepackage[compatible]{algpseudocode} 
\usepackage{upgreek}
\usepackage{xcolor}
\usepackage{pbox}
\usepackage{subcaption}
\usepackage{soul}

\usepackage{enumitem}
\usepackage[version=4]{mhchem}
\usepackage{multirow}
\usepackage{setspace, lipsum}
\usepackage{babel}
\usepackage{blindtext}

\newcommand{\expect}[3]{\mathbb{E}_{#1}\left[#2|#3\right]}
\newcommand{\varian}[3]{\mathrm{var}_{#1}\left[#2|#3\right]}
\newcommand{\real}[1]{\mathbb{R}^{#1}}
\newcommand{\set}[1]{\mathbb{#1}}
\newcommand{\bs}[1]{\boldsymbol{#1}}
\newcommand{\prob}[1]{p\left(#1\right)}
\newcommand{\normdist}[1]{\mathcal{N}\left(#1\right)}

\newcommand{\diag}[1]{\mathrm{diag}\left[#1\right]}
\newcommand{\bst}[2]{\boldsymbol{#1}_{#2}}
\newcommand{\tbst}[2]{\tilde{\boldsymbol{#1}}_{#2}}

\newcommand{\hst}[2]{\hat{\boldsymbol{#1}}_{#2}}
\newcommand{\bbst}[2]{\overline{\boldsymbol{#1}}_{#2}}
\newcommand{\LB}{{\mathbf{u}}_{\mathrm{min}}}
\newcommand{\UB}{{\mathbf{u}}_{\mathrm{max}}}
\newcommand{\ones}{\mathbf{1}}
\newcommand{\eye}{\mathbf{I}}
\newcommand{\bbf}[1]{\bar{\mathbf{#1}}}

\newcommand{\bPhi}{\boldsymbol{\Phi}}
\newcommand{\bphi}{\boldsymbol{\upphi}}
\newcommand{\pos}{\mathbf{p}}
\newcommand{\vel}{\dot{\mathbf{p}}}
\newcommand{\mean}[1]{\boldsymbol{\mu}_{#1}}
\newcommand{\var}[1]{\boldsymbol{\Sigma}_{#1}}
\newcommand{\inv}[1]{#1^{-1}}
\newcommand{\num}[1]{\mathrm{#1}}
\newcommand{\trace}[1]{\mathrm{tr}\left(#1\right)}

\newcommand{\summation}[3]{\sum_{#1}^{#2}#3}
\newcommand{\norm}[1]{\|#1\|_{_2}}
\newcommand{\func}[3]{#1_{#2}\left(#3\right)}

\newcommand{\sbracket}[1]{\left(#1\right)}
\newcommand{\mbracket}[1]{\left[#1\right]}
\newcommand{\lbracket}[1]{\left\{#1\right\}}
\let\oldsubsubsection\subsubsection
\renewcommand{\subsubsection}[1]{\oldsubsubsection{#1}\mbox{}\\}

\author{Cheng-Yu Kuo$^{1*}$, Andreas Schaarschmidt$^{2}$, Yunduan Cui$^{3}$, Tamim Asfour$^{2}$, and Takamitsu Matsubara$^{1}$%
\thanks{$^{1}$Cheng-Yu Kuo and Takamitsu Matsubara are with Graduated School of Science and Technology, Nara Institute of Science and Technology, Nara 630-0192, Japan
        {\tt\footnotesize kuo.cheng-yu.jy5@is.naist.jp; takam-m@is.naist.jp}}%
\thanks{$^{2} $Andreas Schaarschmidt and Tamim Asfour are with Institute of Anthropomatics and Robotics, Karlsruhe Institute of Technology, Karlsruhe 76131, Germany
        {\tt\footnotesize ufdnm@student.kit.edu; asfour@kit.edu}}%
\thanks{$^{3} $Yunduan Cui is with Center for Automotive Electronics, Shenzhen Institutes of Advanced Technology, Chinese Academy of Sciences, Shenzhen 1068, China
        {\tt\footnotesize yd.cui@siat.ac.cn}}%
}

\title{Uncertainty-aware Contact-safe Model-based Reinforcement Learning}

\begin{document}
\maketitle

\begin{abstract}
 This letter presents contact-safe Model-based Reinforcement Learning (MBRL) for robot applications that achieves contact-safe behaviors in the learning process. 
In typical MBRL, we cannot expect the data-driven model to generate accurate and reliable policies to the intended robotic tasks during the learning process due to sample scarcity. 
Operating these unreliable policies in a contact-rich environment could cause damage to the robot and its surroundings. 
To alleviate the risk of causing damage through unexpected intensive physical contacts, we present the contact-safe MBRL that associates the probabilistic Model Predictive Control's (pMPC) control limits with the model uncertainty so that the allowed acceleration of controlled behavior is adjusted according to learning progress. 
Control planning with such uncertainty-aware control limits is formulated as a deterministic MPC problem using a computation-efficient approximated GP dynamics and an approximated inference technique.
Our approach's effectiveness is evaluated through bowl mixing tasks with simulated and real robots, scooping tasks with a real robot as examples of contact-rich manipulation skills. (video: https://youtu.be/sdhHP3NhYi0)
\end{abstract}


\section{Introduction}\label{sec:introduction}
\IEEEPARstart{M}{odel}-based Reinforcement Learning (MBRL) \cite{deisenroth2013ftr,deisenroth2011rss} is attractive in robotics scenarios due to its effectiveness and sample-efficiency. However, we cannot expect the data-driven model to generate reliable control to the intended robotic task during the learning process while sample scarcity. Although applying these unreliable controls are required as 
\emph{exploration}, it could damage the robot and its surroundings in a \emph{contact-rich environment}, for instance, the kitchen. Since safety is the primary consideration, a contact-safe learning process in MBRL needs to be considered.

Several prior works accomplished such a contact-rich manipulation task with Reinforcement Learning (RL): learning a torque profile to pick up objects and open doors  \cite{kalakrishnan2011rsj} and learning billiard strokes and manipulating a box in a chopsticks task with imitation learning refined by RL \cite{pastor2011icra}. A recurrent neural network model performed food-cutting tasks with MBRL \cite{mitsioni2019humanoids}. Although marvelous results, these studies did not consider contact safety during the learning. 

Contact-rich tasks include contact behaviors in which accurate modeling is difficult, and long-horizon forward dynamics prediction which is needed in MBRL, is more challenging \cite{khader2019CoRR}. To alleviate this difficulty, Model Predictive Control (MPC) \cite{mayne2000automatica,cao2017jirs} truncates the prediction horizon and recursively solves problem with a newly observed state that is essentially robust for modeling errors. However, MBRL with MPC may be insufficient to safely achieve contact-rich tasks.

Tactile exploration by touching an unknown object exemplifies the contact-safe behavior of the learning process. Without sufficient knowledge of the object, human beings' fundamental fear and intolerance to uncertainty elicits cautious behaviors when approaching objects to maximize survival \cite{carleton2016jad_2}. Inspired by this concept, the robot would be less likely to have intensive contacts with the environment by moving cautiously during high uncertainty. Therefore, it may be reasonable to assume that associating model uncertainty with the robot's control limits results in contact-safe exploration behaviors.

\begin{figure}
\centering
\includegraphics[width=1\columnwidth]{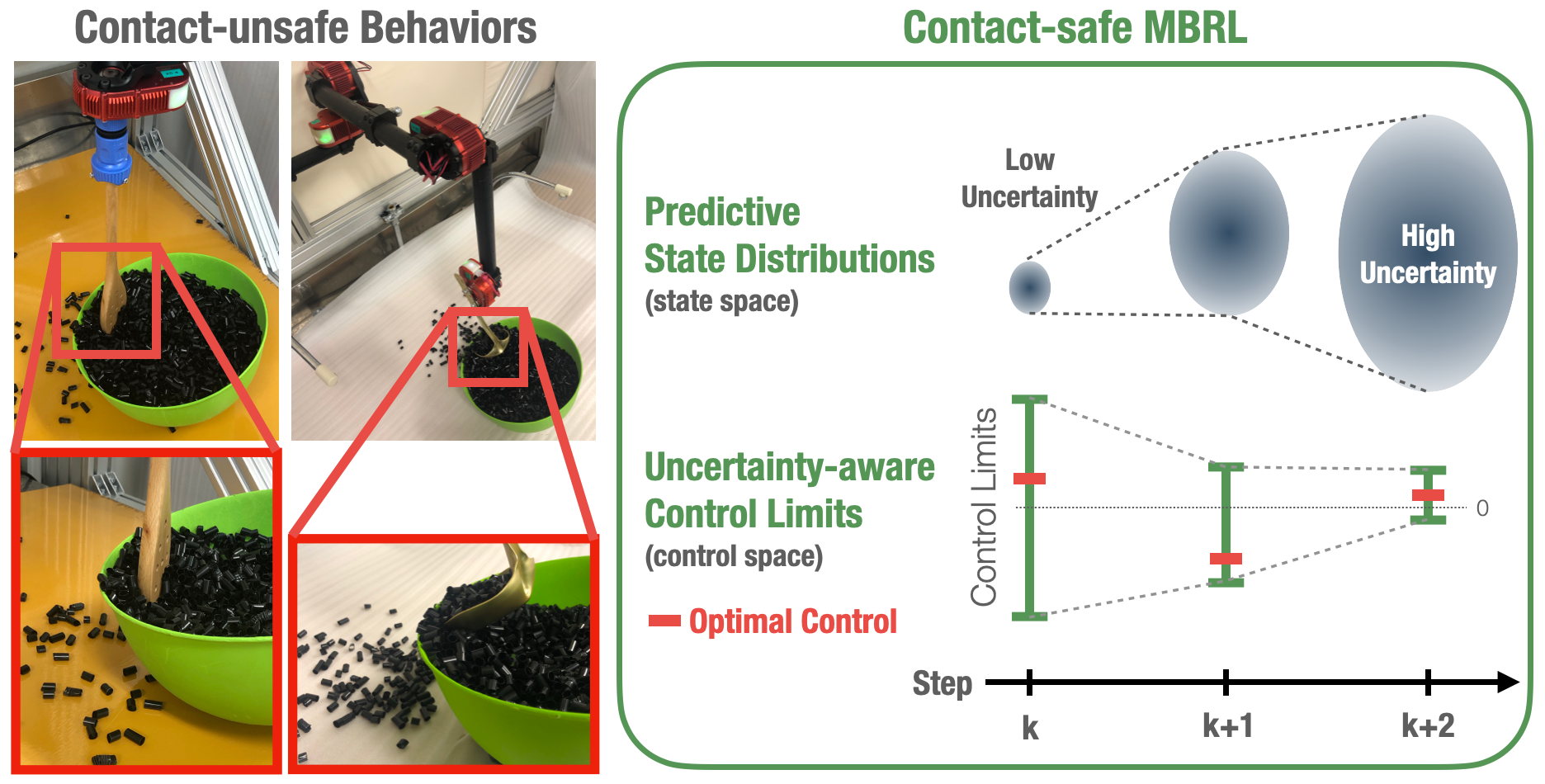}
\caption{(\textbf{Left}) Example of contact-unsafe behavior; (\textbf{Right}) uncertainty-aware control limits in contact-safe MBRL.}
\label{fig:_introduction}
\end{figure}
This letter presents a contact-safe MBRL that reduces the intensity of unexpected contacts during the learning process by promoting the \emph{safety-characteristic}: the agent takes small/gentle actions during high uncertainty. Such safety is achieved by an uncertainty-aware approach that accordingly associates the learning progress, or the model-uncertainty, to handle the control limits of probabilistic-MPC (pMPC) problem (Fig.~\ref{fig:_introduction}). Although prior research handled pMPC control limits using a differentiable squashing function \cite{deisenroth2011icml}, such an approach may result in unreliable predictions near constraint boundaries \cite{schattler2012geo}. A deterministic-reformulated MPC \cite{kamthe2018aistats} was proposed to support dynamic control limits by Pontryagin's Maximum Principle \cite{pontryagin1962pmp}. By following the work \cite{kamthe2018aistats}, we deterministically reformulate a computationally efficient pMPC \cite{kuo2020icra} to implement control planning with such uncertainty-aware control limits.

The following are the contributions of this work:
\begin{enumerate}[leftmargin = *]
    \item We present a contact-safe MBRL that achieves a contact-safe learning process by handling the pMPC control limits with model-uncertainty.
    \item We empirically demonstrate contact-safe behaviors through bowl-mixing tasks with simulated and real robots and scooping tasks with a real robot as examples of contact-rich manipulation skills.
\end{enumerate}

\section{Related Work}\label{sec:related_work}

\subsection{Contact Safety in Contact-rich Manipulations with RL}\label{sec:robot_safeties_in_contact_rich_manipulations with RL}
There are two common approaches to raise contact safety in contact-rich scenarios. The first consists of learning a controller to adjust robot stiffness with RL. For instance, Martín-Martín et al. \cite{martin2019iros} safely performed surface-wiping and door-opening tasks with a model-free, RL-learned variable impedance controller by penalizing forces that exceed a payload. Beltran et al. \cite{beltran2020ral} performed fail-safe ring/peg insertion tasks by learning the gain of the paralleled position/force controller. Wirnshofer et al. \cite{wirnshofer2020rss} achieved a safe peg insertion task with an RL-learned controller that combines a compliance controller with a set of goal-directed policies.

The second approach learns to generate ``safe'' policies to avoid collisions. For instance, Levine et al. \cite{levine2015icra} performed a toy-assembling task by learning a unified control policy from a set of desired trajectories. 
Yamada et al. \cite{yamada2020corl} performed an object-manipulation task in an obstructed environment by switching to a collision-avoidance model-based policy if the model-free RL policy is likely to have a collision.

These approaches focus on contact safety during execution, and the desired behaviors are stated in terms of reward/loss functions. However, such multi-objective optimizations require careful setups to prevent convergence issues \cite{deb2001moo}. In contrast, our approach focuses on contact safety during learning with a single objective loss function for achieving the intended robotic task.

\subsection{Utilizing Uncertainties in Probabilistic Approaches}\label{sec:utilizing_uncertainties_in_probabilisitc_rl_approaches}
Uncertainty in a probabilistic approach provides rich information. Several prior works utilized such information for various purposes. Some utilized uncertainty in a way that encourages agents to expand their exploration coverage \cite{miyamoto2020ral,haarnoja2018icml}. Others plan for a policy that minimizes uncertainty \cite{deisenroth2011icml, cui2020jfr}. Lee et al. \cite{lee2020icra} proposed a guided uncertainty-aware approach that guides the robot to an uncertain area and switches to an RL policy to perform a peg-in-hole task. LaGrassa et al. \cite{lagrassa2020iros} performed a door-opening task by patching the MBRL with an imitation-learned model-free local policy when the MBRL's dynamics model contains high uncertainty.

Unlike the above studies, our method explores a novel application that associates model-uncertainty with the pMPC control limits to adjust the agent's MBRL learning behavior.

\section{Preliminaries}\label{sec:preliminary}
Several pieces of research demonstrated the sample-efficiency of MBRL with Gaussian Processes (GPs) dynamics \cite{deisenroth2011icml,kamthe2018aistats}. However, modeling a standard GPs dynamics largely suffers from computational inefficiency when the sample size is large, resulting in a trade-off between control frequency and the maximum allowed sample size, even though both are equally important in robot applications. Our previous work proposed a computation-efficient MBRL \cite{kuo2020icra} that alleviates the trade-off by using the LGM-FF model, an approximated GPs dynamics. The LGM-FF is a combination of a Linear Gaussian Model (LGM) \cite{williams1998lingp} and \emph{Fastfood} random features \cite{le2013icml} that generate the Fourier features of kernel expressions.

In this work, we utilized the LGM-FF in modeling the system dynamics. Predictions with LGM-FF under uncertain inputs are derived by an analytic moment-matching method \cite{deisenroth2011icml,kuo2020icra} that approximates the future state as a Gaussian distribution. The computation cost of modeling LGM-FF dynamics is $\mathcal{O}\sbracket{NM^3}$ and $\mathcal{O}\sbracket{M^2}$ in exploiting moment-matching under $M$ features and $N$ samples. In contrast, modeling system dynamics and exploiting moment-matching with standard GPs are $\mathcal{O}\sbracket{N^3}$ and $\mathcal{O}\sbracket{N^2}$. The efficiency of \emph{Fastfood} feature selection was proved \cite{le2013icml} and demonstrated  $M\ll N$ in robotic tasks \cite{kuo2020icra}.
\subsection{LGM-FF Dynamics}\label{sec:lgm_ff_dynamics}
Given state input $\bst{x}{t}\in\set{X}\subset\real{D}$ and control signal $\bst{u}{t}\in\set{U}\subset\real{U}$, consider GPs dynamics $f:\set{X}\times\set{U}\to\set{X}$ that represents the system's true underlying dynamics:
\begin{align}
    \begin{split}
        \label{eq:GPs_dynamics}
        \bst{x}{t+1} = f\sbracket{\bst{x}{t}, \bst{u}{t}} + \epsilon
    \end{split},
\end{align}
where $\epsilon\sim\normdist{0, \bs{\Sigma}_n}$ is the system noise with covariance matrix $\bs{\Sigma}_n = \diag{\sigma^2_{n,1},...,\sigma^2_{n,D}}$. For each target dimension $i = 1,...,D$, GPs dynamics $f_i\sbracket{\cdot}$ is modeled by LGM and trained with $\num{N}$ collected samples: $\tbst{X}{} = \mbracket{\tbst{x}{1},...,\tbst{x}{\num{N}}}$ and $\bst{y}{i} = \mbracket{y_{i,1},...,y_{i,\num{N}}}^\top$ that contain individual training input $\tbst{x}{t} := \mbracket{\bst{x}{t}^\top, \bst{u}{t}^\top}{}^\top$ and target $y_{i,t}$ defined as $i$-th element of $\bst{x}{t+1}$. We obtain the predictive distribution given new input $\tbst{x}{*}$:
\begin{align}
    \begin{split}
        \label{eq:LGM_predictive_distribution}
        \prob{f_{i}\sbracket{\tbst{x}{*}}|\tbst{X}{},\bst{y}{i}} \sim \mathcal{N}\sbracket{\bbf{w}_{i}^\top\bphi_{i}^*,\ {\bphi_{i}^*}^\top\inv{\mathbf{A}_{i}}\bphi_{i}^* }
    \end{split},
\end{align}
where $\bphi_{i}\sbracket{\cdot} : \set{X}\times\set{U}\to\real{M}$ is a \emph{Fastfood} feature map with $M$ features, $\bbf{w}_{i}$ is a corresponding weight, and $\bphi_{i}^* := \bphi_{i}\sbracket{\tbst{x}{*}}$. By applying \emph{maximum marginal likelihood estimation} \cite{rasmussen2006gpml,bishop2006prml}, we obtain these optimal parameters:
\begin{align}
    \begin{split}
        \bbf{w}_{i} &= \sigma_{n,i}^{-2}\inv{\mathbf{A}_{i}}\func{\bPhi}{i}{\tbst{X}{}}\bst{y}{i}\quad \quad \quad \quad \ \  \quad \in \real{M\times 1}
    \end{split},\\
    \begin{split}
        \mathbf{A}_{i} &= \sigma_{n,i}^{-2}\func{\bPhi}{i}{\tbst{X}{}}\func{\bPhi}{i}{\tbst{X}{}}^\top + \sigma_{s,i}^{-2}\eye \quad\in \real{M\times M}
    \end{split},
\end{align}
where $\func{\bPhi}{i}{\tbst{X}{}}:= \mbracket{\func{\bphi}{i}{\tbst{x}{1}},...,\func{\bphi}{i}{\tbst{x}{N}}}\in\real{M\times N}$,  $\sigma_{s,i}^{2}$ is the signal variance, and $\eye$ is the identity matrix. 
\subsection{Moment-matching Prediction with LGM-FF}\label{sec:moment_matching_prediction_with_lgm_ff}
We assume that the LGM-FF models for each target dimension are independently trained.
Similar to related works \cite{deisenroth2011icml,cui2020jfr}, assuming probabilistic state $\prob{\bst{x}{t}}=\normdist{\bst{\mu}{t}, \bst{\Sigma}{t}}$ and deterministic control $\bst{u}{t}$, we integrate the LGM-FF dynamics over distribution $\prob{\bst{x}{t}}$ to obtain $\prob{\bst{x}{t+1}} \approx \normdist{\bst{\mu}{t+1}, \bst{\Sigma}{t+1}}$, a Gaussian-approximated predictive distribution  where $i$-th (diagonal) elements of $\bst{\mu}{t+1}$ and $\bst{\Sigma}{t+1}$ are 
\begin{align}
    \begin{split}
        \label{eq:lgm_mm_mu}
        \mu_i &= \ \ \expect{\bst{x}{t}}{\func{f}{i}{\tbst{x}{t}}}{\bst{\mu}{t},\bst{\Sigma}{t}}
        =\bbf{w}_i^\top\mathbf{q}_{i}\ \ \in \set{R}
    \end{split},\\
    \begin{split}
        \label{eq:lgm_mm_sigma}
        \sigma^2_i &= \varian{\bst{x}{t}}{\func{f}{i}{\tbst{x}{t}}}{\bst{\mu}{t},\bst{\Sigma}{t}} \\
        &= \trace{\inv{\mathbf{A}_i}\mathbf{Q}_{i}} + \bbf{w}_i^\top\mathbf{Q}_{i}\bbf{w}_i - \mu_i^2\quad \in\set{R}.
    \end{split}
\end{align}
The analytic solutions of $\mathbf{q}_{i}$ and $\mathbf{Q}_{i}$ were provided in a previous work \cite{kuo2020icra}.

\section{Contact-Safe MBRL}\label{sec:contact_safe_exploration_in_mbrl}
In the following, the ``action'' describes the agent's behavior, and ``control'' denotes the control command sent to the agent.
The contact-safe MBRL's learning process contains $N_{trial}$ trials with $L_{step}$ steps, each of which is for performing task acquisitions. At each step, pMPC plans for an open-loop $H$ step, optimal control sequence $\bst{u}{1},...,\bst{u}{H}$ by recursively exploiting the moment-matching predictions and applying the control $\bst{u}{1}$ to the environment with corresponding sample collecting. Before each trial ends, the LGM-FF dynamics model is updated with all previously collected samples.

During pMPC control planning, state predictions with a data-driven probabilistic dynamics model become more certain based on the sufficiency of the training samples. Thus, we can view the model-uncertainty or the predictive state's variance as a measure that expresses the agent's confidence about its dynamics under the current standpoint. Without sufficient samples, we cannot expect a reliable policy from pMPC with a model that contains high model-uncertainty. However, learning the model requires experiencing the consequences of applying these policies, which may create unexpected yet intensive contacts that cause damage.  

We present a contact-safe MBRL with uncertainty-aware pMPC control limits to reduce the unexpected contact's intensity during the learning process by promoting the \emph{safety-characteristic}: agent takes small/gentle actions during high uncertainty, which is achieved by two \emph{safety-measures}. \emph{safety-measure-A} encourages the agent to take smaller actions by limiting the control during high uncertainty. Since limiting the control also restricts the agent's ability to reduce its velocity, \emph{safety-measure-B} encourages the agent to choose controls that reduce high velocities during high uncertainty.

\subsection{Contact-safe MBRL Framework}
\subsubsection{Deterministic Reformulated System Dynamics}\label{sec:deterministic_system_dynamics_representation}
A computation-efficient pMPC that supports dynamic control limits is achieved by a deterministically reformulated \cite{kamthe2018aistats} LGM-FF dynamics. Predictions with reformulated dynamics $\func{f}{\mathrm{d}}{\cdot}$ follows moment-matching uncertainty propagation (Eqs. (\ref{eq:lgm_mm_mu}-\ref{eq:lgm_mm_sigma})) with expanded state $\bst{s}{t}$ that accommodates both mean $\bst{\mu}{t}$ and variance $\bst{\Sigma}{t}$ as deterministic entries:
\begin{align}
    \begin{split}
        \label{eq:deterministic_dynamics}
        \bst{s}{t+1} = \func{f}{\mathrm{d}}{\tbst{s}{t}} \in \real{2D+U}\to\real{2D},
    \end{split}
\end{align}
where $\bst{s}{t} := \mbracket{\bst{\mu}{t}^\top, \ones^\top\bst{\Sigma}{t}}^\top \in \set{S}\subset\real{2D}$, and $\tbst{s}{t} := \mbracket{\bst{s}{t}^\top, \bst{u}{t}^\top}^\top \in\real{2D+U}$. $\ones$ denotes a column vector of ones. 

In the following applications, since robots are velocity-controlled  with discrete acceleration control signals, we define the entries of deterministic reformulated state $\bst{s}{}$:
\begin{align}
    \begin{split}
        \label{eq:state_definition}
        \bst{\mu}{} = \mbracket{\bst{\mu}{\pos}^\top,\bst{\mu}{\vel}^\top}^\top,\ 
        \bst{\Sigma}{} = \diag{\ones^\top\bst{\Sigma}{\pos}, \ones^\top\bst{\Sigma}{\vel}}
    \end{split},
\end{align}
where $\pos$ and $\vel$ are the actuators' position velocity; $\mean{\pos}$, $\mean{\vel}$, $\var{\pos}$ and $\var{\vel}$ are their mean vector and covariance matrices.
\begin{figure}
\centering
\includegraphics[width=1\columnwidth]{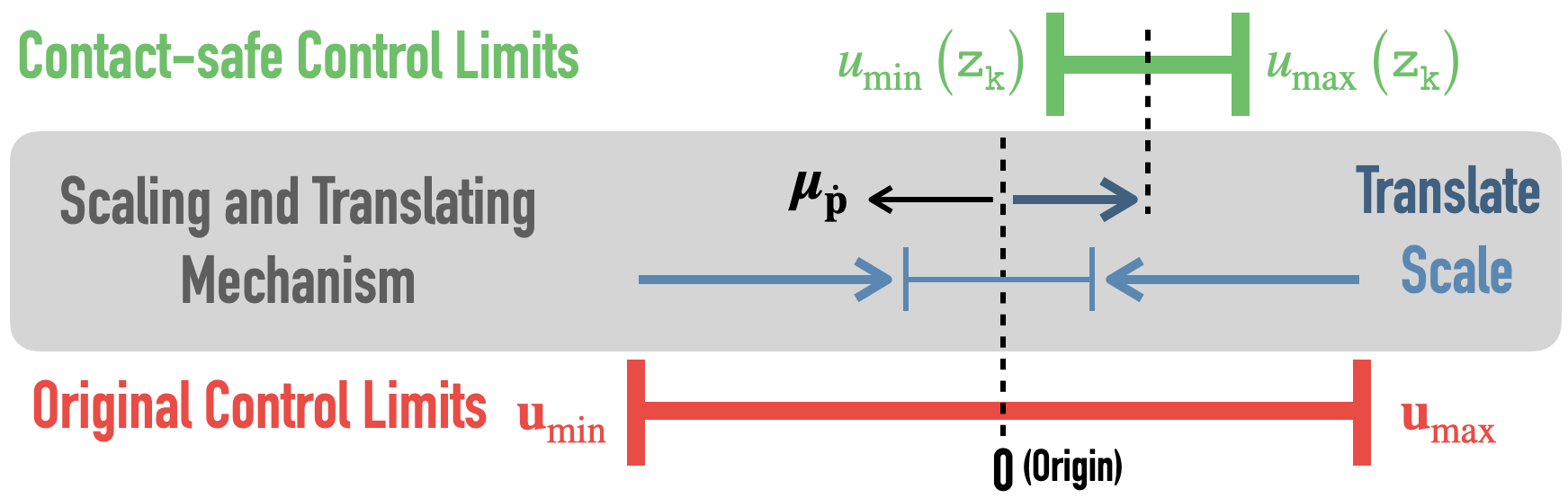}
\caption{Scaling and translating mechanisms modify control limits during high model-uncertainty. Translating the limits encourages the agent to reduce its velocity $\bst{\mu}{\vel}$.}
\label{fig:_contact_safe_control_limits}
\end{figure}
\subsubsection{pMPC with Uncertainty-aware Control Limits}\label{sec:contact_safe_mbrl}
We utilize the deterministic reformulated dynamics (Eq.~(\ref{eq:deterministic_dynamics})) in a novel way by handling the pMPC control limits with the model-uncertainty which exists in predictive state $\hst{s}{}$.
At time step $t$, for all $H$ steps, the pMPC control limits are adjusted by functions $\func{u}{\mathrm{min}}{\hst{s}{k}},\func{u}{\mathrm{max}}{\hst{s}{k}}$:
\begin{equation}
\small
\left.
\begin{alignedat}{4}
        \label{eq:pMPC_dynamic_control_limits}
        &\text{minimize}\quad   &&\func{\mathcal{L}}{}{\bst{s}{t}} &&=     &&\summation{k = 2}{H+1}{\ell\sbracket{\hst{s}{k}}} \\
        &\text{subject to}      &&\hst{s}{k+1}  &&=     &&\func{f}{\mathrm{d}}{\bbst{s}{k}},  \bbst{s}{k} = \mbracket{\hst{s}{k}^\top, \hst{u}{k}^\top}^\top,\hst{s}{1} = \bst{s}{t}\\
        &                       &&\hst{s}{k}    &&\in   &&\set{S},\ k=1,...,H+1\\
        &                       &&\hst{u}{k}    &&\in   &&\mbracket{ \func{u}{\mathrm{min}}{\hst{s}{k}},\func{u}{\mathrm{max}}{\hst{s}{k}}},\ k=1,...,H
\end{alignedat}
\right\}
\end{equation}
where $\ell:\set{S}\to\real{}$ is the immediate loss. 

\subsubsection{Control Limits Function with Adjustable Awareness}\label{sec:adjustable_uncertainty_aware_control_limit_function}
We propose linear control limits that are handled by scaling and translating mechanisms that associate model-uncertainty to achieve the \emph{safety-characteristic}. During high uncertainty, scaling down the control limits fulfills \emph{safety-measure-A}, and translating the control limits to reduce the agent's velocity fulfills \emph{safety-measure-B}. Therefore, we design functions $\func{u}{\mathrm{min}}{\hst{s}{k}},\func{u}{\mathrm{max}}{\hst{s}{k}}:\set{S}\to\set{U}$:
\begin{alignat}{3}
    \label{eq:upper_bound}
    &\func{u}{\mathrm{max}}{\hst{s}{k}} &&= \func{K}{s}{\hst{s}{k}}\UB &&- \ \func{K}{t}{\hst{s}{k}},\\
    \label{eq:lower_bound}
    &\func{u}{\mathrm{min}}{\hst{s}{k}} &&=  \underbrace{\func{K}{s}{\hst{s}{k}}}_{\mathrm{scaling}}\LB &&- \underbrace{\func{K}{t}{\hst{s}{k}}}_{\mathrm{translating}},
\end{alignat}
where $\func{K}{s}{\cdot}: \set{S}\to\set{R}$ and $\func{K}{t}{\cdot}: \set{S}\to\set{R}^U$ associate state $\hst{s}{k}$ to modify the feasible control limits of agent $\set{U}\in\mbracket{\LB,\UB}$. Both $\LB,\UB\in\mathbb{U}$ are identically adjusted. Fig.~\ref{fig:_contact_safe_control_limits} illustrates the scaling and translating mechanisms. 

The following properties for $\func{K}{s}{\cdot}$ fulfill \emph{safety-measure-A}:
\begin{enumerate}[leftmargin = 25pt,label= (A\arabic*)]
    \item decreases during high uncertainty.
    \item has a pre-defined minimum value within $(0, 1]$ to prevent control limits with zero or negative range.
    \item has a maximum value of $1.0$ during low uncertainty to return to the baseline: $\hst{u}{k}\in\mbracket{\LB,\UB}$.
    \item has a parameter for tuning the uncertainty sensitivity.
\end{enumerate}

The following properties for $\func{K}{t}{\cdot}$ fulfill \emph{safety-measure-B}:
\begin{enumerate}[leftmargin = 25pt,label= (B\arabic*)]
    \item increases with velocity during high uncertainty.
    \item has a minimum value of $0$ during low uncertainty to return to the baseline: $\hst{u}{k}\in\mbracket{\LB,\UB}$.
    \item has a parameter for tuning the uncertainty sensitivity.
    \item has a parameter for tuning the velocity sensitivity.
\end{enumerate}

With the above in mind, we design $\func{K}{s}{\cdot}$ and $\func{K}{t}{\cdot}$ with exponential decay functions, which is a simple way to create mappings of non-linear decay or growth (flipping the decay function) within the range of $\mbracket{0, 1}$ under $\set{R}_{\ge 0}$ inputs:
\begin{align}
    \begin{split}
        \label{eq:scaling_factor_function}
        \func{K}{s}{\hst{s}{}} &= 
        \underbrace{(1-\beta_s)}_{(\mathrm{A3})}
        \overbrace{\exp(-
        \underbrace{\alpha_s}_{(\mathrm{A4})}
        \|\bst{\Sigma}{\pos}\|_{_2})}^{(\mathrm{A1})}
        + \underbrace{\beta_s}_{(\mathrm{A2})}
    \end{split},\\
    \begin{split}
        \label{eq:translating_factor_function}
        \func{K}{t}{\hst{s}{}} &=
        \overbrace{\underbrace{(1 - \exp(-
        \underbrace{\alpha_t}_{(\mathrm{B3})}
        \|\bst{\Sigma}{\pos}\|_{_2}))}_{(\mathrm{B2})}
        \underbrace{\gamma_t}_{(\mathrm{B4})}
        \bst{\mu}{\vel}}^{(\mathrm{B1})}
    \end{split},
\end{align}
where $\alpha_s, \alpha_t\in\set{R}_{\ge 0}$ are the tunable uncertainty-awareness parameters. $\gamma_t\in\set{R}_{\ge 0}$ is the velocity sensitivity, which is usually set as the ratio of the control's feasible upper bound to the agent's max velocity to prevent over translation. $\beta_s$ is the minimum scaling value. For a consistent scale of uncertainty, only position variance $\bst{\Sigma}{\pos}$ is measured to estimate the uncertainty at the current state. 

\begin{figure}
\centering
\includegraphics[width=1\columnwidth]{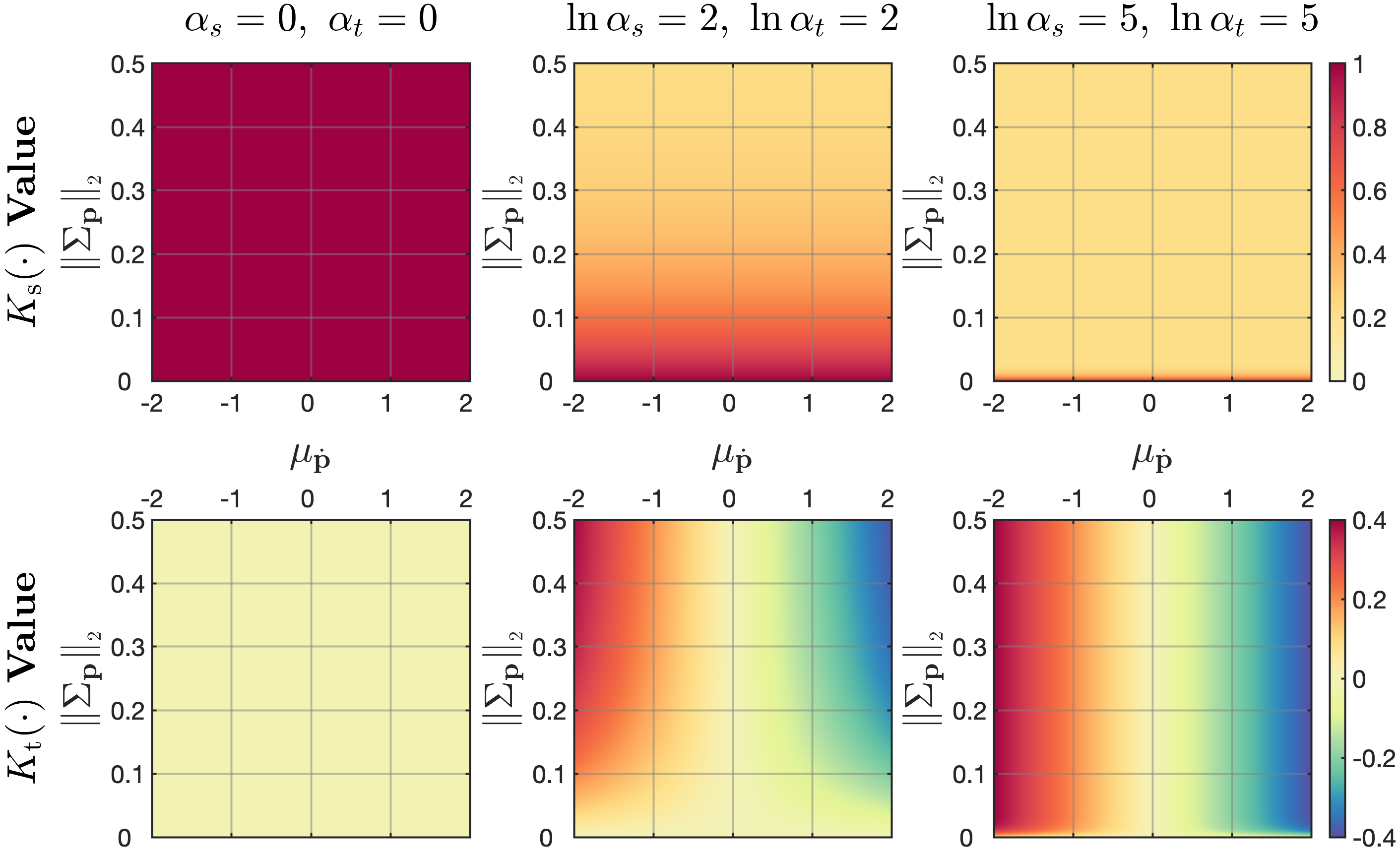}
\caption{Relationship between values of $\func{K}{s}{\cdot}, \func{K}{t}{\cdot}$ and state $\bst{s}{}$ under different $\alpha_s, \alpha_t$ settings. Color indicates corresponding values.}
\label{fig:_kvalues_imagesc}
\end{figure}

We adjust the agent's learning behavior by changing uncertainty-awareness parameters $\alpha_s,\alpha_t$. For instance, we return to the standard contact-unsafe MBRL with $\alpha_s,\alpha_t = 0$ or a contact-safe MBRL with excessive uncertainty-awareness $\ln\alpha_s, \ln\alpha_t = 10$. Assuming a single dimensional $\bst{\mu}{\vel}$, Fig.~\ref{fig:_kvalues_imagesc} exemplifies the relationship between states $\bst{s}{}$ and $\func{K}{s}{\cdot},\func{K}{t}{\cdot}$ under $\beta_s, \gamma_t = 0.2$. Note that the translating is only active while $\bst{\mu}{\vel}\neq 0$ under a certain level of uncertainty.

\subsubsection{Ahead pMPC Planning for Delay Compensation}\label{sec:ahead_prediction_for_control_delay_compensation}
\begin{figure}
\centering
\includegraphics[width=1\columnwidth]{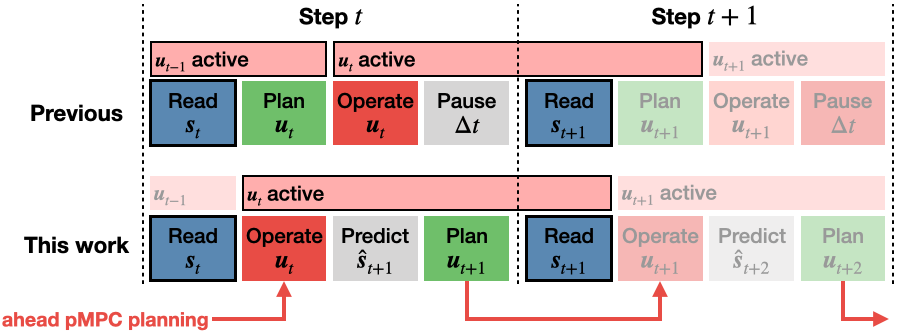}
\caption{Control delay compensation with one-step ahead pMPC planning. }
\label{fig:_ahead_prediction}
\end{figure}

Our previous works's MBRL framework suffers from inconsistent active control signals between the states of each time step \cite{kuo2020icra}. Similar to the parallel MPC workflow \cite{cui2020jfr} and asynchronous control \cite{yang2019corl}, we introduce the ahead pMPC planning to alleviate the inconsistency. At time step $t$, the ahead pMPC planning plans for $n$ step ahead control signal $\bst{u}{t+n}$ based on a predicted state $\hst{s}{t+n}$ that obtained by recursively applying state predictions via Eq.~(\ref{eq:deterministic_dynamics}). Since the control signal $\bst{u}{t}$ was previously planned, the ahead pMPC planning compensates for the control delay and improves the consistency between $\bst{s}{t}$ and $\bst{s}{t+1}$ (Fig.~\ref{fig:_ahead_prediction}). Also, removing the pausing improves the control frequency. Above all, we could estimate the uncertainty level of the following state $\bst{s}{t+1}$ with the predicted state $\hst{s}{t+1}$ for handling the control limits.

\begin{algorithm}
  \SetKwFunction{reset}{ResetPosition}
  \SetKwFunction{observe}{ObserveState}
  \SetKwFunction{operate}{Operate}
  \SetKwFunction{train}{trainLGM}
  \textbf{Initial input} \\
  \quad Number of trials: $N_{trial}$. step length: $L_{step}$.\\
  \quad Empty sample set: $(\tbst{s}{}, \bst{Y}{})$. pMPC horizon: $H$.\\
  \quad Long-term loss: $\mathcal{L}$. Immediate loss: $\ell$.\\
  \quad Awareness parameters: $\alpha_s, \alpha_t$.\\
  \# \textbf{Generate LGM-FF model}\\
  \quad$f_{\mathrm{d}}\leftarrow$ \train{$\tbst{s}{},\bst{Y}{}$}\\
  \# \textbf{MBRL process}\\
  \For{$n = 1,2,...,N_{trial}$}{
    \reset{}\\
    \# \textbf{One-step Ahead pMPC planning with Eq.~(\ref{eq:pMPC_dynamic_control_limits})} \\
    \quad$\bst{s}{1}$ = \observe{}\\
    \quad $\bst{u}{}^\star$ = $\arg\min_{\bst{u}{}}\mathcal{L}\sbracket{\bst{s}{1}}$\\
    \For{$j = 1, 2, ..., L_{step}$}{
      \quad$\bst{s}{j}$ = \observe{}\\
      \quad\operate{$\bst{u}{}^\star(1)$}\\
      \quad$\tbst{s}{j} = \mbracket{\bst{s}{j}, \bst{u}{}^\star\sbracket{1}}$\\
      \# \textbf{One-step Ahead pMPC planning with Eq.~(\ref{eq:pMPC_dynamic_control_limits})} \\
      \quad$\hst{s}{j+1}$ =  $\func{f}{\mathrm{d}}{\tbst{s}{j}}$  \\
      \quad $\bst{u}{}^\star$ = $\arg\min_{\bst{u}{}}\mathcal{L}\sbracket{\hst{s}{j+1}}$\\
      \quad$\bst{y}{j}$ = \observe{}\\
      \quad$\tbst{s}{}= \lbracket{\tbst{s}{}, \tbst{s}{j}}, \bst{Y}{}=\lbracket{\bst{Y}{}, \bst{y}{j}}$
    }
    \# \textbf{Update LGM-FF model}\\
    \quad$f_{\mathrm{d}}\leftarrow$ \train{$\tbst{s}{},\bst{Y}{}$}
  }
  \caption{Contact-safe MBRL}
  \label{alg:MBRL_sdpmpc}
\end{algorithm}
\subsubsection{Summary}\label{sec:summary}
Algorithm~\ref{alg:MBRL_sdpmpc} summarizes the contact-safe MBRL with the one-step ahead pMPC planning. First, the deterministic reformulated dynamics model allows us to associate the pMPC control limits with the model-uncertainty in a principled way. Next we designed the uncertainty-aware control limits function by exponential decay functions to meet the \emph{safety-characteristic}. Last, we added ahead pMPC planning to compensate for the control delay.

Utilizing an uncertainty-aware approach offers two benefits: 
\begin{enumerate}[leftmargin = *]
    \item The limitation relaxes adaptively and allows bigger actions while the knowledge (or samples) becomes sufficient.
    \item The agent can react to a sudden uncertainty increase while entering a novel area and adjusts its behavior accordingly.
\end{enumerate}

\begin{figure}
\centering
\includegraphics[width=1\columnwidth]{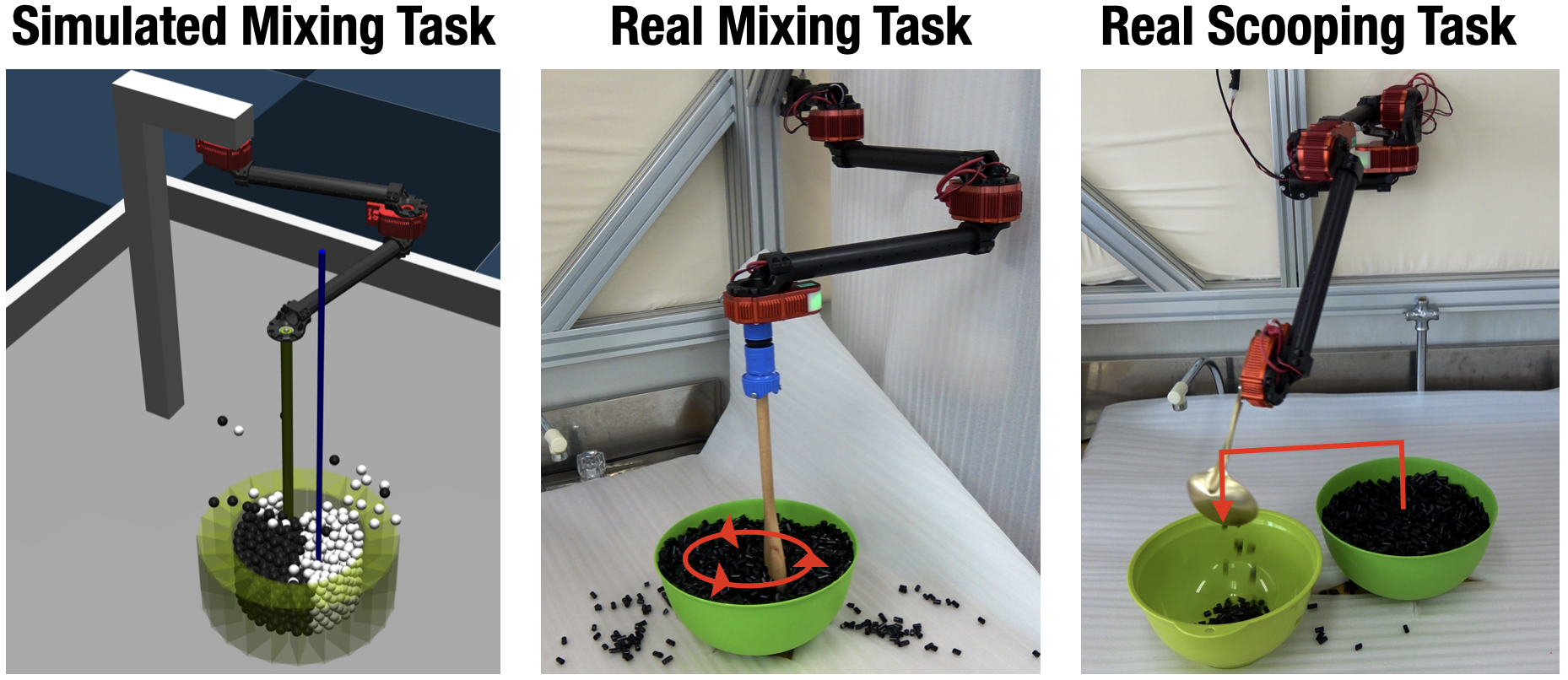}
\caption{(\textbf{Left}) Mixing task with simulated 2-DoF HEBI arm: (Blue stick: reference target position at current step) (\textbf{Middle}) Mixing task with real 2-DoF HEBI arm. (\textbf{Right}) Scooping task with real 4-DoF arm.}
\label{fig:_experiment_env}
\end{figure}
\section{Experimental Evaluation}\label{sec:experiment}
\subsection{Overall Definition}
Mixing and scooping are commonly performed during cooking \cite{bollini2013}. Since acquiring such tasks from scratch may cause danger in a contact-rich environment, we selected them to justify our approach's effectiveness. We conducted both simulated and real-robot experiments with HEBI robotics' hardware (Fig.~\ref{fig:_experiment_env}). The simulator emulates a real-robot mixing setup to evaluate our approach's effectiveness under various uncertainty-awareness settings. Real-robot experiments verified the real-world potential of our approach.

Due to the actuators' periodic behavior, we defined the position and velocity of a $\mathrm{k}$-Degree-of-Freedom (DoF) robot in Eq.~(\ref{eq:state_definition}): $\pos:= [\sin\bst{\Theta}{}^\top, \cos\bst{\Theta}{}^\top]^\top,\vel := \dot{\Theta}$, the $\Theta = [\theta_1,...,\theta_{\mathrm{k}}]^\top$ denotes the rotational position. For smooth movements, robots are velocity-controlled with discrete acceleration control signal $\bst{u}{}= \Delta\vel/\Delta t$.

The objective of our experiments is learning the robot's dynamics from scratch while performing the intended task by tracking the provided hints: the world-space reference end-effector trajectories for each task. To emphasize the intensity of unexpected contacts during the learning process with force/torque sensors, the reference trajectories of the tasks do not contain stiff contacts.
\begin{figure}
\centering
\includegraphics[width=1\columnwidth]{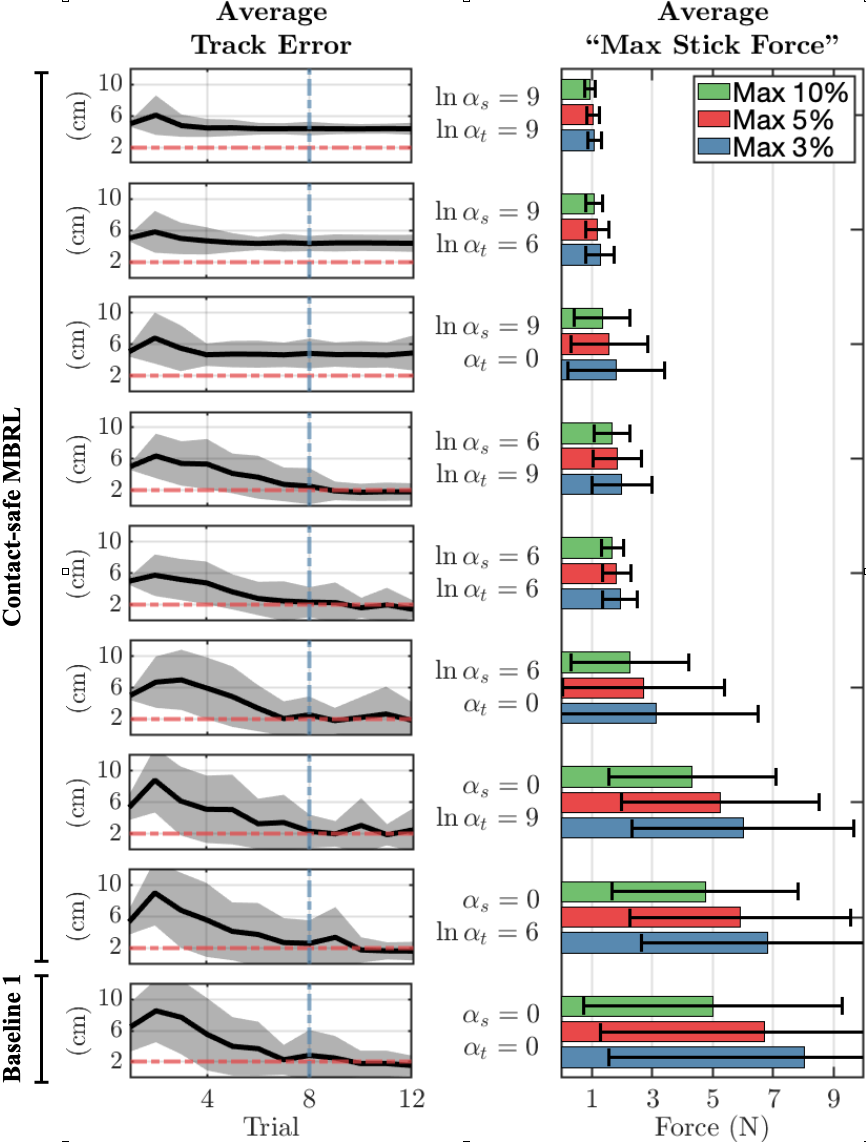}
\caption{(\textbf{Left}) Average tracking error and (\textbf{right}) measured ``max stick force" over 20 experiments under each setting. (Red dashed line indicates $2$ cm to distinguish task acquisition status. Blue dashed line indicates eighth trial to judge learning efficiency.)}
\label{fig:_20trials_summary}
\end{figure}
\begin{figure}
\centering
\includegraphics[width=1\columnwidth]{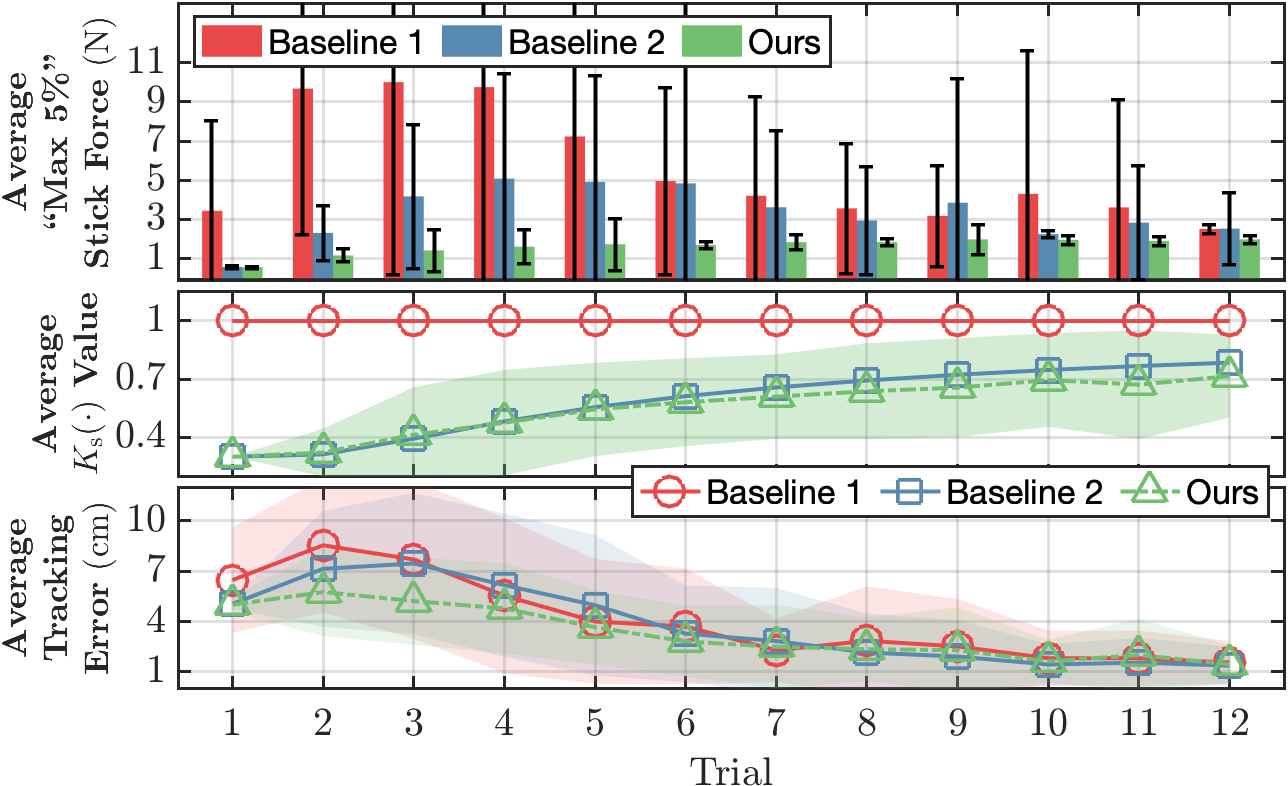}
\caption{(\textbf{Upper}) Average ``max 5\% stick force:'' (\textbf{Middle}) Average $\func{K}{s}{\cdot}$ value, inversely proportional to uncertainty level. (\textbf{Bottom}) Average tracking error to show learning efficiency. All values are averaged over 20 experiments and color fills illustrate 95\% confidence interval. 
(\textbf{Baseline 1}: contact-unsafe MBRL. \textbf{Baseline 2}: $\lbracket{\ln\alpha_s,\ln\alpha_t=6}$ contact-safe MBRL with global uncertainty: replace position uncertainty $\|\bst{\Sigma}{p}\|_2$ with inverted sample size $N^{-1}$. \textbf{Ours}: $\lbracket{\ln\alpha_s,\ln\alpha_t=6}$ contact-safe MBRL.)}
\label{fig:_20trials_2_baseline}
\end{figure}
For better confirmation of the proposed method, loss function $\ell$ does not contain a regularization term for $\bst{u}{}$. The contact-safe MBRL is responsible for all the behavior changes. The loss function is designed to fulfill the objective:
\begin{align}
    \begin{split}
        \label{eq:loss_function_definition}
        \ell\sbracket{\bst{s}{t}} = \underbrace{\mathrm{k}_s\norm{\bst{p}{t}^{ref} - \bst{p}{t}}}_{\text{position loss}} + \underbrace{\mathrm{k}_o\norm{\bst{o}{t}^{ref} - \bst{o}{t}}}_{\text{orientation loss}},
    \end{split}
\end{align}
where $\mathrm{k}_s, \mathrm{k}_o \in\set{R}_{\ge 0}$ are corresponding weights and $\bst{p}{t},\bst{o}{t}$ are the robot's end-effector position and orientation in the world-space at time step $t$ that track references $\bst{p}{t}^{ref}, \bst{o}{t}^{ref}$. We obtain $\bst{p}{t},\bst{o}{t}$ by forward kinematics with joint positions calculated by an Euler equation: $\hst{\Theta}{} = \mathrm{Re}\mbracket{-i\sbracket{\bst{\mu}{\cos{\bst{\Theta}{}}} + i\bst{\mu}{\sin{\bst{\Theta}{}}} }}$.

The tracking error is defined as $\norm{\bst{p}{t}^{ref} - \bst{p}{t}}$. All the experiments share these parameters: number of features $M = 65$; pMPC horizon $H = 3$; all entries of $\LB$ and $\UB$ are $-0.2$ and $0.2$ rad/s; $\beta_s = 0.3$; $\gamma_t = 0.2$; $\mathrm{k}_s,\mathrm{k}_o = 1.0$.

\subsection{Simulation Experiment}\label{sec:simulation_experiment}
\subsubsection{Simulator Setup}\label{sec:simulator_setup}
We emulate a real particle-mixing task environment in the Mujoco simulator (Fig.~\ref{fig:_experiment_env}). The two-DoF is $0.65$-m-long in total with a $0.02$-m-diameter cylinder-shaped stick attached to its end-effector. The stick is $0.5$-m-long and does not contact the bowl's bottom during mixing. A force sensor is attached to the stick to measure contact forces. A $0.24$-m-diameter bowl is fixed below the arm and filled with $900$ $0.03$-m-diameter particles whose total mass is $9$ kg. The reference trajectory follows a $0.1$-m-diameter circulating pattern with a cycling period of five seconds. The learning process contains $12$ trials and $100$ steps with $0.1$-second step interval.

\subsubsection{Overall Result}\label{sec:statistic_result}
This section demonstrates the effectiveness of our proposed contact-safe MBRL by comparing two baselines:
\begin{itemize} [leftmargin=*]
    \item \emph{Baseline-1}: the standard contact-unsafe MBRL (uncertainty-aware control limits disabled $\alpha_s, \alpha_t=0$).
    \item \emph{Baseline-2}: contact-safe MBRL with globally evaluated uncertainty by sample size ($\|\bst{\Sigma}{\mathbf{p}}\|_2$ replaced by $N^{-1}$).
\end{itemize} 

We conducted $20$ experiments with both baselines and our approach under various uncertainty-awareness settings. The learning efficiency is demonstrated by averaging the tracking error at each trial. We averaged the collected top $\mbracket{3\%,5\%,10\%}$ measured stick forces to demonstrate how effectively our approach reduced the contact intensity.

\textbf{Contact-safe vs. contact-unsafe:} Fig.~\ref{fig:_20trials_summary} shows the effectiveness of our approach by comparing \emph{baseline-1} and our approach under the following combinations of uncertainty-awareness settings: $\mbracket{\alpha_s = 0, \ln\alpha_s = 6, \ln\alpha_s = 9}$ and $\mbracket{\alpha_t = 0, \ln\alpha_t = 6, \ln\alpha_t = 9}$. Here $\alpha_s = 0$ or $\alpha_t = 0$ disables the corresponding modifying mechanism. In addition to combinations with $\ln\alpha_s = 9$ that failed the acquisition due to overly conservative behavior, other combinations show similar learning efficiencies with \emph{baseline-1} for a task that was acquired within $8$ trials. By comparing the max measured forces, our approach significantly reduced the contact intensity, especially with setting $\lbracket{\ln\alpha_s, \ln\alpha_t = 6}$.

\textbf{Uncertainties for contact-safe MBRL:} Fig.~\ref{fig:_20trials_2_baseline} shows the benefit of utilizing the model-uncertainty of our approach compared with \emph{baseline-2} under the best uncertainty-awareness setting we experienced: $\lbracket{\ln\alpha_s,\ln\alpha_t = 6}$. \emph{Baseline-1} is also provided. Despite similar learning efficiency across two baselines and our approach, the \emph{baseline-2} and our approach result in significantly lower averaged contact forces throughout the learning process. However, contact forces observed in \emph{baseline-2} has a higher standard deviation than our approach ($8$-th, $9$-th, and $11$-th trial), which indicates that \emph{baseline-2} has a higher possibility to encounter intensive contacts. This is because \emph{baseline-2} globally evaluates the uncertainty by sample size, which cannot handle novel situations after collecting a certain amount of samples. In contrast, our approach adaptively evaluates the uncertainty by model-uncertainty. As a result, the $\func{K}{s}{\cdot}$ value of our approach holds high variances throughout $15$ trials and resulting in the lowest overall contact intensity.

\begin{figure*}
\centering
\includegraphics[width=2\columnwidth]{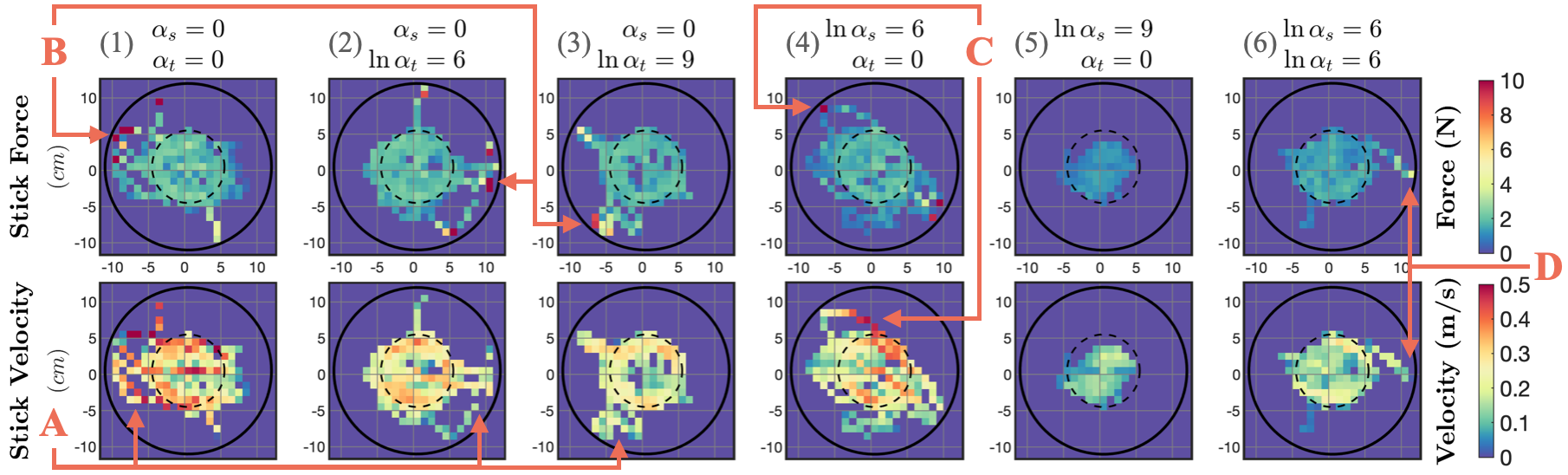}
\caption{Stick force (\textbf{upper}) and velocity (\textbf{lower}) analysis under various awareness settings through $12$ trials: Figure's top view shows bowl boundary (black solid lines) and reference mixing trajectory (black dashed lines). Color differences indicate max value observed at each location in bowl. (\textbf{1}): standard contact-unsafe MBRL with contact-safe MBRL with (\textbf{2}, \textbf{3}): scaling disabled, (\textbf{4}, \textbf{5}): translating disabled, and (\textbf{6}): best combination setting. (A) Translating reduces velocity of entering novel area. (B) Intensive contacts are disabled during scaling. (C) Intensive contacts while exploring novel area caused by high velocities when translating is disabled. (D) Contacts are explored safely when both scaling and translating are enabled.}
\label{fig:_force_velocity_mapping}
\end{figure*}
\begin{figure}
\centering
\includegraphics[width=1\columnwidth]{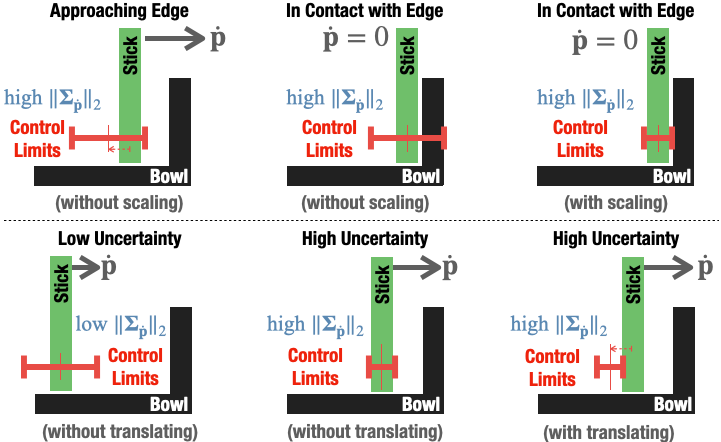}
\caption{Changes in control limits (\textbf{upper}) with/without scaling and (\textbf{lower}) with/without translating.}
\label{fig:_no_scaling_translating}
\end{figure}

\subsubsection{Individual Results}\label{sec:individual_result}
We selected the individual results of the following settings to demonstrate the scaling and translating mechanisms' roles (Fig.~\ref{fig:_force_velocity_mapping}): (\textbf{1}): the standard contact-unsafe MBRL $\lbracket{\alpha_s,\alpha_t = 0}$; (\textbf{2}, \textbf{3}): scaling disabled under translating settings $\ln\alpha_t = 6,9$; (\textbf{4}, \textbf{5}): translating disabled under scaling settings $\ln\alpha_s = 6,9$; (\textbf{6}): combination settings $\lbracket{\ln\alpha_s, \ln\alpha_t = 6}$.

Figure~\ref{fig:_force_velocity_mapping}-(A) shows that increasing $\alpha_t$ effectively reduces the velocity of the explored novel area. However, without scaling, intensive contacts occur despite the lower velocities (Fig.~\ref{fig:_force_velocity_mapping}-(B)). Touching the bowl's edge significantly reduces the velocity that disables the translating mechanism (Fig.~\ref{fig:_no_scaling_translating}: upper). Fig.~\ref{fig:_force_velocity_mapping}-(C) emphasizes the essence of \emph{safety-measure-B}. Without translating, intensive contacts occurred because the robot's capability to reduce its velocity in a novel area is limited by the scaling (Fig.~\ref{fig:_no_scaling_translating}: lower).  The setting (\textbf{5}) shows overly conservative behavior whose task acquisition failed. By enabling both scaling and translating, the robot acquired the task at much lower contact forces with its capability to safely explore the contacts (Fig.~\ref{fig:_force_velocity_mapping}-(D). 

Although the setting (\textbf{6}) shows smaller exploration coverage, our approach does not discourage the robot from exploring novel areas; it encourages safe exploration. With our approach, the robot can safely expand its knowledge coverage during the learning process.

\subsection{Real-Robot Experiment}\label{sec:real_robot_experiment}
Our real-robot experiment contains two tasks: a mixing task that verifies the real-world performance from simulation results and a scooping task that demonstrates the potential of the proposed method on a larger scale (Fig.~\ref{fig:_experiment_env}). Both tasks are conducted with a contact-unsafe MBRL $\lbracket{\alpha_s,\alpha_t = 0}$ and a contact-safe MBRL with setting $\lbracket{\ln\alpha_s ,\ln\alpha_t = 5}$. We compared the learning efficiency, measured joint torques, and the post-environment of the mixing task acquisitions' results. For the scooping task, we only show the environment differences after the task was acquired as an example of applying the contact-safe MBRL on a larger scale.

\subsubsection{Mixing Task with Two-DoF Arm Configuration}\label{sec:mixing_task}
\textbf{Task setup:} 
The configuration of the real-robot's mixing task resembles that of the simulation setup, where a stick is attached to the end-effector without contacting the bottom of the bowl during mixing. The $0.24$-m-diameter metal bowl is filled with pieces of cut drinking-straws to imitate a cooking ingredient. The reference trajectory follows a circulating pattern with a $0.1$-m-diameter and three-second periods. The learning process contains $15$ trials and $100$ steps, each of which has a $0.1$-second duration.

\begin{figure}
\centering
\includegraphics[width=1\columnwidth]{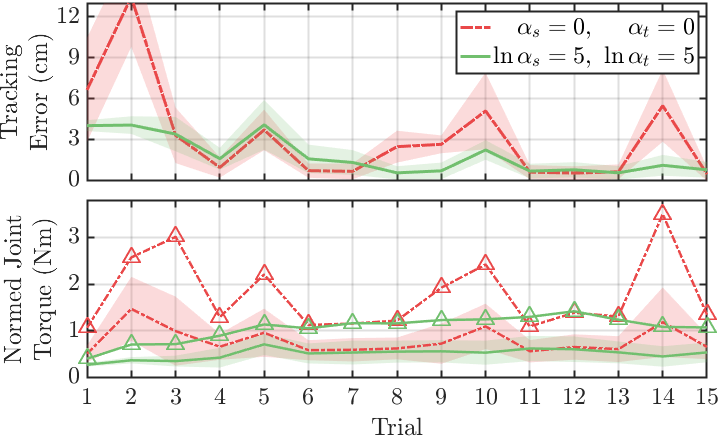}
\caption{(\textbf{Upper}) Average tracking error and (\textbf{lower}) average and max ($\triangle$) measured torques through trials (mixing task)}
\label{fig:_2dof_real_error_torque}
\end{figure}

\textbf{Results:} 
Since the measured joint torques from the actuator sensor increase during intensive contacts, we collected joint torques during the MBRL process to evaluate contact intensity. In Fig.~\ref{fig:_2dof_real_error_torque}, the contact-safe MBRL under setting $\lbracket{\ln\alpha_s, \ln\alpha_t = 5}$ shows similar results to the simulations where the contact intensities are significantly reduced and a similar learning efficiency with the standard contact-unsafe MBRL $\lbracket{\alpha_s, \alpha_t = 0}$. Also, the contact-safe MBRL's less intensive behavior also scattered fewer straws in the environment after the learning process, Fig.~\ref{fig:_environment_result_and_success}.

\subsubsection{Scooping Task with Four-DoF Arm Configuration}\label{sec:scooping_task}
\textbf{Task setup:} The scooping task is comprised of a four-DoF arm with a spoon attached to the end-effector and two bowls: one filled with pieces of straws and one empty. Along with the velocity control, the arm is gravity-compensated during the learning process. The reference trajectory is a series of spoon's positions and orientations that scoop and transport the pieces of straws in world-space.
This experiment demonstrates the effectiveness of our approach for changing learning behavior in a large workspace.
The learning process contains $40$ trials and $65$ steps, each of which has $0.2$-second intervals.

\begin{figure}
\centering
\includegraphics[width=1\columnwidth]{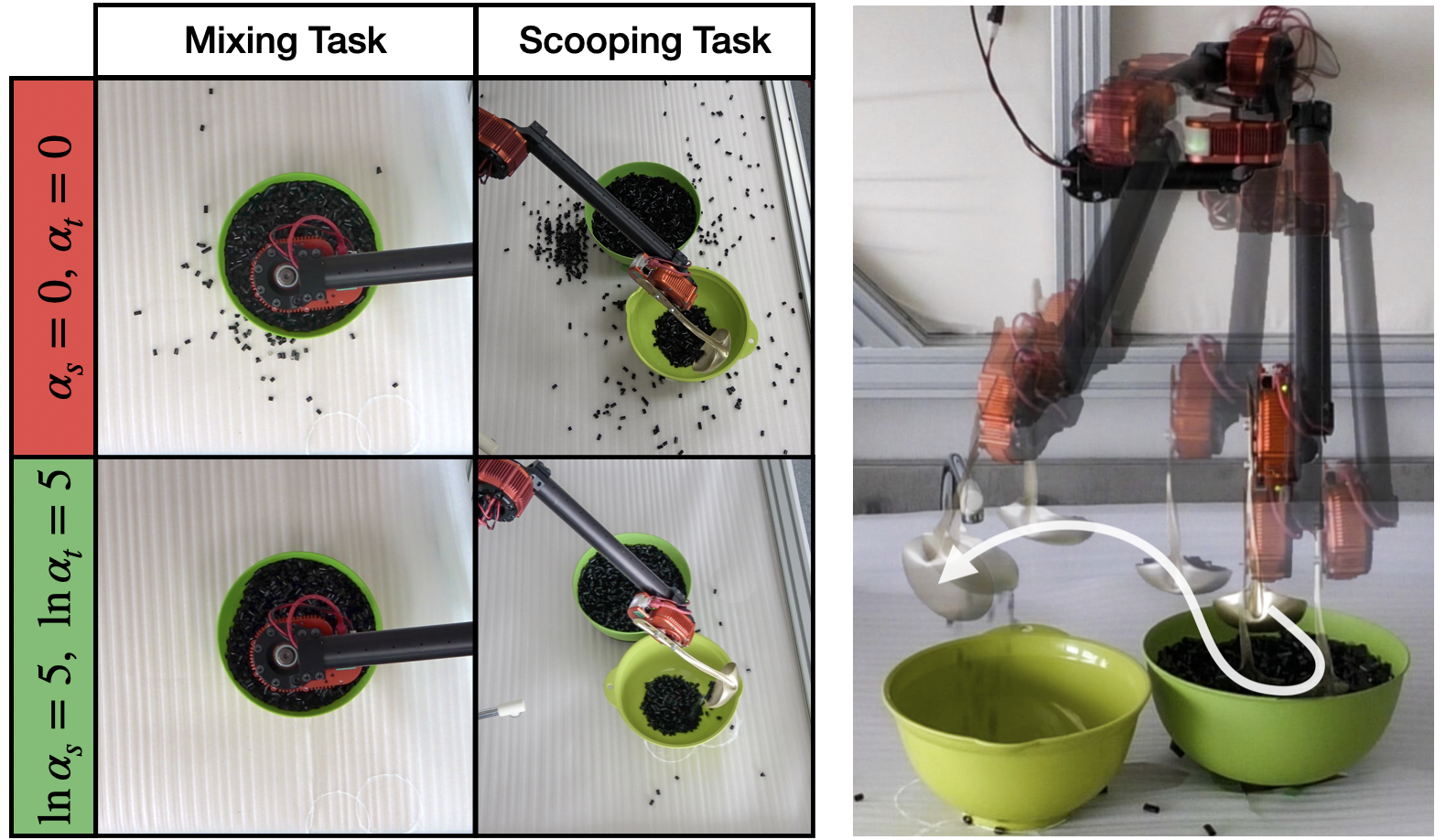}
\caption{(\textbf{Left}) Environment differences after 15 trials of mixing tasks and six successful scooping tasks. (\textbf{Right}) Learned scooping trajectory with our approach.}
\label{fig:_environment_result_and_success}
\end{figure}

\textbf{Results:} After subtracting the torque that was applied as gravity compensation, both contact-unsafe MBRL and our approach achieved over six successful scoops within $40$ trials, resulting in a very similar pattern that reduced the measured torques shown in Fig.~\ref{fig:_2dof_real_error_torque}. Due to space limitations, here we only show the environment differences after the learning process. After six successful scoops that transported the straws to the target bowl, judging by the number of straws scattered in the environment, the contact-safe MBRL significantly reduced the intensity of the learning process (Fig.~\ref{fig:_environment_result_and_success}). This result verifies the behavior-changing ability of our contact-safe MBRL on a larger scale. The learning processes of these two settings are shown in the attached video.
\section{Discussions And Future Work}\label{sec:discussions_and_future_work}
This work presents a contact-safe MBRL that reduces the contact intensities during learning processes by utilizing uncertainty information. Given that ideal behavior varies among operators, we recommend that users select uncertainty-awareness parameters by starting with overly conservative settings and gradually reducing their values until the agent reaches ideal results. Although our experimental results show that the contact-safe MBRL has similar learning efficiency to standard MBRL, our experiment's kitchen tasks have a much-limited workspace. If the task would have a more complex and larger workspace, our approach could penalize the learning efficiency for exploring cautiously. 

Despite our proposal's effectiveness, we discuss several future works that we will investigate to improve this work:
\begin{enumerate}[leftmargin = *]
    \item Integrating compensations in the contact-safe MBRL: This work's control limits have an origin of zero, where additional compensations are needed to perform tasks that require continuous stabilizing, e.g., against gravity.
    \item Integrating measures for contact safety during executing.
    \item Using multiple-output GPs dynamics \cite{alvarez11a2011jmlr}: In this work, we modeled the dynamics of each state's dimension individually. However, the state's entries could be much correlated in complex contact-rich tasks, and capturing that may be beneficial. 
\end{enumerate}

\bibliography{library}
\bibliographystyle{ieeetr}
\end{document}